\documentclass[conference]{IEEEtran}
\IEEEoverridecommandlockouts
\usepackage{cite}
\usepackage{amsmath,amssymb,amsfonts}
\usepackage{algorithmic}
\usepackage{graphicx}
\usepackage{textcomp}
\usepackage{xcolor}

\usepackage[T1]{fontenc}
\usepackage[utf8]{inputenc}

\usepackage{booktabs}
\usepackage{algorithm2e}
\usepackage{graphicx}
\usepackage{multirow}

\usepackage{pgfplots} 
\pgfplotsset{compat=newest}
\usepgfplotslibrary{groupplots}
\usepgfplotslibrary{dateplot}
\usepackage{tikz}
\usepackage{xlop}

\usepackage{xlop}
\usetikzlibrary{math}
\usepackage{datatool}
\usepackage{pgfplotstable}
\pgfplotsset{compat=1.8}

\usepackage{array}
\usepackage{colortbl}
\def\BibTeX{{\rm B\kern-.05em{\sc i\kern-.025em b}\kern-.08em
    T\kern-.1667em\lower.7ex\hbox{E}\kern-.125emX}}

\begin{document}

\title{Semantic Similarity To Improve Question Understanding in a Virtual Patient}

\author{\IEEEauthorblockN{ Fréjus A. A. Laleye}
\IEEEauthorblockA{\textit{CEA, LIST, Laboratoire d'Analyse Sémantique Texte et Image}\\
Gif-sur-Yvette, F-91191 \\
frejus.laleye@cea.fr}
\and
\IEEEauthorblockN{ Antona Blanié}
\IEEEauthorblockA{\textit{Laboratoire de Formationpar la Simulation } \\ \textit{ et l’Image en Médecine et en Santé (LabForSIMS),}\\
 Faculté de Médecine Paris-Sud, 94275 Le Kremlin Bicêtre \\
antonia.blanie@aphp.fr}
\and
\IEEEauthorblockN{ Antoine Brouquet}
\IEEEauthorblockA{\textit{Laboratoire de Formationpar la Simulation } \\ \textit{ et l’Image en Médecine et en Santé (LabForSIMS),}\\
 Faculté de Médecine Paris-Sud, 94275 Le Kremlin Bicêtre \\
antoine.brouquet@aphp.fr}
\and
\IEEEauthorblockN{ Dan Benhamou}
\IEEEauthorblockA{\textit{Laboratoire de Formationpar la Simulation } \\ \textit{ et l’Image en Médecine et en Santé (LabForSIMS),}\\
 Faculté de Médecine Paris-Sud, 94275 Le Kremlin Bicêtre \\
dan.behnamou@aphp.fr}
\and
\IEEEauthorblockN{ Gaël de Chalendar}
\IEEEauthorblockA{\textit{CEA, LIST, Laboratoire d'Analyse Sémantique Texte et Image}\\
Gif-sur-Yvette, F-91191 \\
gael.de-chalendar@cea.fr}
}

%

\maketitle

\begin{abstract}
In medicine, a communicating virtual patient or doctor allows students to train in medical diagnosis and develop skills to conduct a medical consultation. In this paper, we describe a  conversational  virtual  standardized  patient  system to allow medical  students  to  simulate  a  diagnosis strategy of an abdominal surgical emergency. We exploited the semantic properties captured by distributed word representations to search for similar questions in the virtual patient dialogue system. We created two dialogue systems that were evaluated on datasets collected during tests with students. The first system based on hand-crafted rules obtains $92.29\%$ as $F1$-score on the studied clinical case while the second system that combines rules and semantic similarity achieves $94.88\%$. It represents an error reduction of $9.70\%$ as compared to the rules-only-based system.

\end{abstract}


\maketitle



\section{Introduction}
\label{sec:introduction}

The medical diagnosis practice is traditionally bedside taught. Theoretical courses are supplemented by internships in hospital services. The medical student observes the practice of doctors and interns and practices himself under their control. This type of learning has the disadvantage to confront immediately the medical student with complex situations without practical training (technical and human) beforehand. At the same time, he must manage relations with people in pain and the mobilization of complex and incomplete knowledge. It therefore seems useful to be able to train before confronting his first patients. But for this to be realistic, it can not be done with peers who would play the role of patients. Realism would be insufficient. It is sometimes done with actors playing a learnt case. This is called a Standardized Patient \cite{hubal_virtual_2000}. But it presents several limitations, like ``actor training and availability, reproducibility, changing evaluation criteria, and implementation costs''.

Advances in virtual reality allow diving, cheaply, the student in a realistic and pedagogically controlled environment. Virtual Standardized Patient (VSP) technology is used to manage the basics of a standardized dialogue between a medical student and a virtual patient. It is  generally two-dimensional avatars on a computer display that serve to help learners maneuver through clinical situations or perform a task, such as the proper technique for taking a patient's medical history \cite{christine2019}. 
For an educational purpose in a real learning environment, VSP becomes unusable in the sense that the slightest deviation from the intended scenario or the lexicon causes the communication to fail.

In this paper, we demonstrate how to exploit the performance of two different neural classification methods, semantic similarity techniques using sentence embeddings and rules pattern matching to improve the process of language understanding in a dialogue system in order to produce an realistic VSP that can be used by medical students. Another contribution in this work is the dataset construction of doctor-patient dialogues that will be made publicly available. A paper describing in details its building method, content and access point is already submitted. 

\section{Related work}
\label{sec:related_work}

Recent advances in speech recognition technologies, natural language processing and artificial intelligence have led to the emergence of conversational agents in different domains of life such as health, finance, education, etc.
In the health domain, the last decades have seen the expansion of virtual patients or doctors that are used to interact with humans in real clinical scenarios simulation for training purposes, education or medical evaluation.
For medical training, students play the physician  role to diagnose virtual patient's pain and prescribe appropriate treatments by extracting symptoms in validated clinical and educational interview scenarios
\cite{safari2014}. Virtual patients take different forms according to educational objectives and already exist in virtual reality simulators for endoscopy training \cite{harpham2015}. They are also used, to teach the techniques of oral examination to interns in emergency medicine \cite{mcgrath2015}, to enable nurses to develop acute care skills such as assessment and management of clinical deterioration \cite{mcgrath2015}, and trainees to practice clinical reasoning techniques \cite{close2015,kleinert2015}.
Several medical programs have already incorporated virtual simulations including dummies to assess the competence and confidence of learners to lead clinical situations \cite{taglieri2017}. 
Different evaluations show that the use of virtual patients provides additional practices for learners outside laboratory work and improves their performance on actual clinical cases. Smith and colleagues prove that virtual patient technology can improve learner performance on clinical consultation issues \cite{smith2017}.

Much research is underway to improve the effectiveness of virtual conversational agents in fulfilling the ever-increasing needs of healthcare professionals. All of these projects aim to increase and improve the interaction between patients and physicians \cite{bioulac2018,campillos2016,campillos2017}. 
This interaction, in virtual simulations using conversational agents, is characterized by:
\begin{itemize}
    \item the type of technology (platform hosting the conversational agent) namely smartphones \cite{miner2016,ireland2016}, laptops or desktops \cite{tanaka2017,philip2017} and multimodal platforms \cite{lucas2017};
    \item the dialogue management strategy namely the finite state strategy where the dialogue is a sequence of predefined steps \cite{tanaka2017,philip2017,lucas2017}; the content-based strategy where the dialog flow is not predefined but depends on the content provided by the user \cite{ireland2016}; the agent-based strategy where the dialogue is between two agents capable of reasoning  \cite{miner2016};
    \item the initiative in the dialogue where it is either the user who initiates the conversation \cite{miner2016}, either the virtual agent that leads the conversation \cite{tanaka2017,philip2017,lucas2017}, or both \cite{fitzpatrick2017,ireland2016}.
\end{itemize}

State-of-the-art methods have been developed to respond to the challenge of question understanding and interpreting of virtual patient dialogue systems. Similar to our approach, the method described in \cite{jin2017} combined machine learning techniques and pattern matching for Question Interpretation. The authors demonstrated the value of combining a hand-authored pattern matching system with word and character-based convolutional neural network for improving question identification in a virtual patient dialogue system. Machine learning techniques are widely used in the medical field to improve medical treatment or diagnosis. They allow to build embedded models in AI systems that are used by the physicians in order to analyze patient data and to make predictions on possible outcome of a treatment or to provide additional data for a diagnosis or a prognosis \cite{sebastian2019}. They play increasingly important roles in pre-procedural planning for complex surgeries and interventions and in processing of the historical records of emerging surgical techniques. This to improve medical education and allow students to practice by the simulation. In  \cite{kellen2019}, the authors developed a virtual standardized patient system that can understand, respond, categorize, and assess student performance in gathering information during a typical medical history, thus enabling students to practice their history-taking skills and receive immediate feedback. 

The aims of the current work are: 1) the development of a medical conversational agent (virtual patient) to interact with the physician using speech; 2) demonstrate the utility of using words embeddings to find similar questions in a virtual patient system.

\section{Virtual Standardized Patient System}
\label{sec:vsps}

The virtual agent built in this work is a 42-year-old woman, a kindergarten teacher with a medical history, who consults for abdominal pain. The pedagogical challenge is the teaching of the best practices of diagnostic strategy of an abdominal surgical emergency. The simulation consists of the care of the patient by the medical learner. This is a serious game in virtual reality in which the student, in a virtual environment, dialogs with the patient, examines her, requests exams and produces a diagnosis according to the information provided by the patient and her examination. The virtual environment was built with a set of 3D models showing an emergency room and the character of the sick woman. The 3D models and patient motions and behaviors were built with Unity 3D game engine by our industrial partner. The different motions and behaviors are based on the dialogue acts, the scenario and instructions given by the emergency physician (the medical student). The student can move around the environment by teleporting and can communicate with it via an HTC{\textregistered} controller. Our VSP dialogues with students via a speech recognizer and a speech synthesis module. 

In this paper, we describe the dialogue system integrated in the Virtual Standardized Patient and the strategy of combining machine learning methods and pattern matching for the good understanding of the open student's questions. 

\section{Dataset}
\label{sec:dataset}

\begin{figure}[htb]
\centering
\begin{tikzpicture}[yscale=0.02,xscale=0.9]

\DTLforeach*{questionsextension}{\xA=x, \yA=before}{%
    \tikzmath{
        int \Y;
        \Y = \yA / 10;
    }
    \draw [] (\xA,\Y) node[above,xshift=-2mm,scale=0.7]{\yA};
}
\DTLforeach*{questionsextension}{\xA=x, \yA=before}{%
    \tikzmath{
        int \Y;
        \Y = \yA / 10;
    }
    \draw[line width=2mm,color=red!50,xshift=-2mm] plot[ycomb] coordinates {(\xA,\Y)};
}

\DTLforeach*{questionsextension}{\xA=x, \yA=after}{
    \tikzmath{
        int \Y;
        \Y = \yA / 15;
    }
    \draw [] (\xA,\Y) node[above,xshift=2mm,scale=0.2]{\yA*10};
}
\DTLforeach*{questionsextension}{\xA=x, \yA=after}{%
    \tikzmath{
        int \Y;
        \Y = \yA / 15 ;
    }
    \draw[line width=4mm,color=blue!50,xshift=2mm] plot[ycomb] coordinates {(\xA,\Y)};
}

\draw (0,0) grid[xstep=8,ystep=50] (8,200);

\DTLforeach*{questionsextension}{\xA=x, \l=label}{%
    \draw (\xA,0) node [below left,rotate=45,scale=0.7] {\l};
}
\end{tikzpicture}
\caption{Number of questions by categories before (scaled up 10 times for visibility) and after data augmentation}
\label{fig.dataset_size1}
\end{figure}
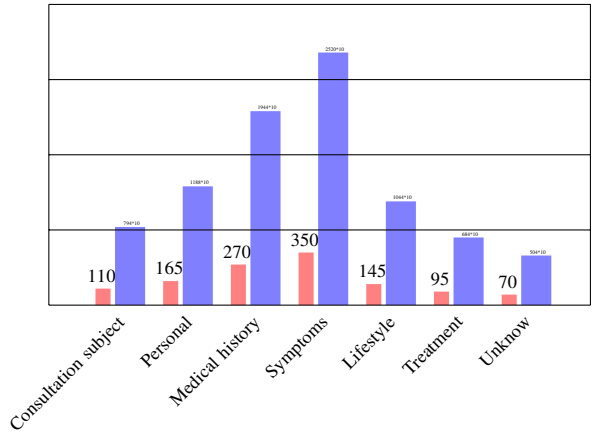

We collected data from multiple sources to form different datasets that are used to process question categorization and semantic distance computation tasks (Section \ref{sec:ssbs}). These tasks are the components of the approach proposed in this paper for improving the understanding mechanism of a virtual patient. To build the different models, we first collected the French dialogue subtitles of \textit{Chicago Med} television series and the data used in 
\cite{campillos2017} for a task of automatic doctor-patient question classification. The collected questions were manually annotated according to the question categories of a surgical consultation for abdominal pain listed in Table \ref{tab:tab1}. These categories have been chosen by the medicine doctors authors of this work. We consider them as fairly consensual in medicine. The questions are used to build the knowledge base of the semantic similarity based system. We then enrich all of the collected data by augmenting the questions using the concepts (synonym classes) created while writing dialog rules with ChatScript, the rule-based chatbot system we use (Section \ref{sec:pms}). As an illustration, the question \textit{Do you have trouble urinating?} and the concept containing the words \textit{urinate} and \textit{pee} will give the following questions in the augmented corpus: \textit{Do you have trouble urinating?} and \textit{Have you struggled to pee?}. The concepts, equivalent to WordNet synonymy classes, were written manually according to the studied domain and without any knowledge from an external database.
In total, $86780$ questions are obtained for the categorization and semantic distance calculation tasks.
Figure \ref{fig.dataset_size1} presents the number of questions by categories before and after data augmentation. As questions from Chicago Med dialogs concerned different clinical cases, we have adapted the answers to make them coherent with our own clinical case.

\begin{filecontents*}{questions_extension.txt}
x,label,before,after
1,Consultation subject,110,794
2,Personal,165,1188
3,Medical history,270,1944
4,Symptoms,350,2520
5,Lifestyle,145,1044
6,Treatment,95,684
7,Unknow,70,504
\end{filecontents*}
\DTLloaddb[noheader=false]{questionsextension}{questions_extension.txt}

To form the test dataset, we deployed the rule-based system in a medical school to assess the virtual patient's understanding of questions asked by 25 interns with prior knowledge about the purpose of the medical consultation and the question categories to ask in this case. The responses provided by the patient are then manually corrected by the medical experts. The test sessions with the interns also allowed to adjust and add dialogue rules to have a broad coverage of the clinical case. This dataset contains about $400$ well-annotated question-answer pairs to evaluate the dialogue systems. Furthermore, we extracted from \textit{Chicago Med} subtitles a set of about $250$ questions containing both questions of a consultation for abdominal pain and others that move away from this clinical case. The purpose of this dataset is to evaluate the virtual patient's ability to understand as many questions as possible with the same number of dialogue rules. Each dataset integrates different formulations of the same question.

\section{Pattern Matching system}
\label{sec:pms}

The pattern matching system represents the core of our Standardized Virtual Patient dialogue manager. It uses the ChatScript software \cite{chatscript2013} and includes the question understanding engine and the response generator.
ChatScript was primarily used in this work as a rule-based engine because of its powerful pattern matching mechanism aiming at detecting meaning that is well suited to doctor-patient interactions whose scope is well defined. Furthermore, its open nature and the fact that it is developed in C++ helps its integration in our software stack.
It was used to write a set of pattern-based rules for dialogue management. 
A rule is defined by a pattern that is associated with an action to be triggered.
About 1500 pattern-based rules have been written to cover the clinical case and the patient-student interactions.
The rules were written for identification and understanding of general (name, age) and specific to a medical consultation (medical history, current treatment) questions. These are questions that should lead the student to make a diagnosis and to propose a treatment. We identified seven categories of questions encountered during a surgical emergency consultation. Thus, our patterns are arranged into these seven categories: consultation subject, personal questions, medical history, symptoms, lifestyle, treatment and unknown. The latter includes any matter not belonging to other categories. Table \ref{tab:tab1} presents some question topics encountered in the targeted clinical case for each category. 

\definecolor{Gray}{gray}{0.9}
\newcolumntype{g}{>{\columncolor{Gray}}c}

\begin{table*}[htb]
\centering
\begin{tabular}{@{}c|g|c|g|c|g|c@{}}
\toprule
\begin{tabular}[c]{@{}l@{}}Consultation \\ subject \end{tabular} & Personal & Medical history& Symptoms & Lifestyle & Treatment & Unknow \\ \midrule
The Goal & Age & Family history & Sickness history & Addictions & Type of treatment &  Everything else \\
 & Weight & Past medical history & Changes/evolutions & Pets & Method &  \\
 & Housing & Past surgical history & Location &  & Date and period &  \\ 
 & Job & Allergies & Timing/chronologie &  & Observation &  \\
 & Children & Medications taken & Quantification/severity &  &  &  \\ \midrule
\end{tabular}

\caption{Example of question topics for each category}
\label{tab:tab1}
\end{table*}

Our dialogue management system based on pattern rules is conceived to make clinical examination of the virtual patient more dynamic by taking into account variations in question formulation. So, we have pattern rules to manage, among others, variation in the wording of the same intention, terminological variation of medical concepts, memory in dialogue, abbreviated questions understanding and conversational markers.
Table \ref{tab:tab2} lists some examples of the types of interactions managed by our dialog system. This management has enabled to prevent the student's weariness during the medical consultation and allows to take into account several types of question structures. This leads the student to obtain a larger part of information necessary for the diagnosis. 

\begin{table}[htb]
\centering
\begin{tabular}{|@{}l|l@{}|}
\hline
Interaction types & Examples   \\
\hline
Wording variation & \textit{\begin{tabular}[c]{@{}l@{}}Doctor: What symptoms do you have?\\ Patient: I feel stomach pain\\ \hline Doctor: What is happening?\\ Patient: I have stomach pain\end{tabular}}   \\
\hline
Terminological variation & \textit{\begin{tabular}[c]{@{}l@{}}Doctor: Do you have any medical history?\\ Patient: No doctor\\ \hline Doctor: Have you had health problems \\ in the past?\\ Patient: No doctor\end{tabular}}   \\
\hline
Memory & \textit{\begin{tabular}[c]{@{}l@{}}Doctor: did you take medication?\\ Patient: yes I took two paracetamol\\ \hline Doctor: how much did you take?\\ Patient: I said two\end{tabular}}  \\
\hline
Abbreviated question & \textit{\begin{tabular}[c]{@{}l@{}}Doctor: do you have allergies ?\\ Patient: I'm allergic to pollen\\ \hline Doctor: Do you have a treatment for this?\\ Patient: No treatment\end{tabular}}  \\
\hline
Conversational markers & \textit{\begin{tabular}[c]{@{}l@{}}Doctor: What is happening?\\ Patient: I have stomach pain\\ \hline Doctor: where exactly ?\\ Patient: above the sternum\end{tabular}} \\
\hline
\end{tabular}

\caption{Examples of interaction types}
\label{tab:tab2}
\end{table}

A question asked by the student is first preprocessed with the LIMA linguistic analyzer \cite{lima2010} coupled with ChatScript. LIMA performs the question preprocessing (stop word removal, lemmatization, part of speech tagging) and returns the normalized form obtained as input to ChatScript. Input is then analyzed by ChatScript for specific topics, concepts, phrases, or keywords.

\section{Semantic Similarity Based System}
\label{sec:ssbs}

The second approach developed to make the virtual patient dialogue system acceptable and usable is a system that combines question classification and semantic similarity computation.
The assumption of this approach is that ensuring a wide coverage of student's questions will improve understanding performance of the rule-based system. It is based on sentence embeddings for computing a similarity measure between two questions $q_i$ and $q_j$ ($q_i$ is the student's question and $q_j$ a question from the knowledge base of the virtual patient).
It consists of a set of neural sentence classifiers that take as input the pre-trained word embeddings \cite{fasttext2016} of a question by assigning it a category and a semantic distance comparator between the vector of the question and the set of questions from the identified category. 

\subsection{Classification Model}
\label{sec:cms}

The purpose of our classification task in this approach is the categorization of the questions for which the virtual patient must provide an answer to the student.
The challenge is to build efficient classification models to make the dialogue management efficient and robust. 
For that, we trained two different classification models using different learning methods.
We built a first question classifier using Convolution Neural Networks with an architecture similar to \cite{kim-2014-convolutional}.  Then, we built a second linear classifier based on FastText continuous word representation with rank constraint \cite{fasttextclassif2017}. For each question category, a binary classifier is built leading to assigning class $1$ to questions in a given category and $0$ to those not belonging to that category.

For each model, we trained five submodels with different splits of data in $80\%$ training and $20\%$ validation. We then combined the submodels using the majority voting method on the submodel outputs.
All the questions in each dataset were preprocessed. We first labeled the questions with the seven categories. Then, we proceeded to the normalization of the questions (spellcheck, punctuation removing and lemmatization) and to text cleaning by removing stop words, numerical values and punctuation. 

\subsubsection{Training}
\label{sec:mt}

The questions are represented in a sequence of words used as inputs to the submodels trained separately on the data. We used pre-trained word embeddings for words from all datasets. The embeddings we used are the default wiki-news FastText \cite{fasttext2016}. To allow batch processing of our data, each sentence in the dataset is extended to the maximum sentence length, which turns out to be 50. This maximum length is justified by the short and precise form of the questions generally asked by the doctor during a consultation. 

We used $10$-fold cross-validation for training and validation with a ratio $90/10$ for splitting the dataset. The submodels were trained on different training data obtained from the random generation at different places. After validation, we obtained the best-performing parameters (Table \ref{tab:parameters}) that are used for the evaluation of models embedded in the dialogue system.

\begin{table*}[htb]
\centering
\begin{tabular}{@{}cccccc@{}}
\toprule
\textbf{Kernel size} & \textbf{Kernel number} & \textbf{Embedding dimension} & \textbf{Dropout} & \textbf{learning rate} & \textbf{Optimizer} \\ \midrule
3 to 5 & 300 & 300 & 0.4 & 3.0 & Adamdelta \\ \bottomrule
\end{tabular}
\caption{Hyperparameter values}
\label{tab:parameters}
\end{table*}

\subsubsection{Voting}
\label{sec:voting}

To obtain the better performances of each of the submodels and a reduction of variance, we used a decision-making strategy based on both the majority voting of the submodels and a weighted linear interpolation of the classification model results. The class predicted by a binary classifier is the one that obtained the most votes from the submodels. 
The output of a classifier is given by the binary classifier having predicted a class with the highest probability. The decision-making process is summarized in algorithm \ref{algo:a1} where  $\hat{Y_c}$ is the output of $c$-th binary classifier, $\hat{y_d}$ is the output of $d$-th submodel, $\hat{Y_e}$ is the output of $e$-th classification method, $\alpha_e$, called validation coefficient of the binary classifier, is the accuracy  of $e$-th classifier trained on validation data and $\hat{Y_r} $ is the final output. At the end of the process, the predicted class is $argmax(\hat{Y_r})$. 


\begin{algorithm}
\SetAlgoLined
\DontPrintSemicolon
    \ForEach{ classification method }{
        \ForEach{binary classifier}{
            $\hat{Y_c} \longleftarrow  max(\hat{y_d})$
        }
        $\hat{Y_e} \longleftarrow  max(\hat{Y_c})$
    }
    $\hat{Y_r} \longleftarrow  softmax(\sum_e \alpha_e \hat{Y_e})$

    \Return $argmax(\hat{Y_r})$

\caption{Algorithm for combination of submodel outputs}
\label{algo:a1}
\end{algorithm}

\subsection{Classification Results}
\label{sec:cr}

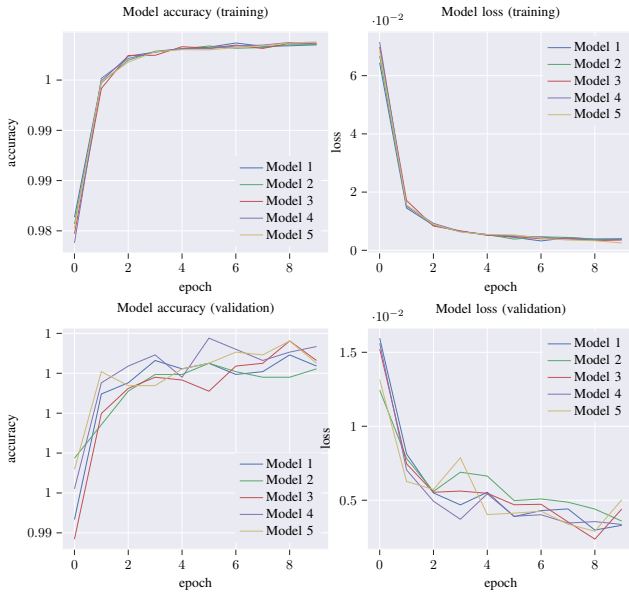
\begin{figure}[htb]
\centering
\resizebox{8.5cm}{!}{%
\begin{tikzpicture}

\definecolor{color0}{rgb}{0.917647058823529,0.917647058823529,0.949019607843137}
\definecolor{color1}{rgb}{0.298039215686275,0.447058823529412,0.690196078431373}
\definecolor{color2}{rgb}{0.333333333333333,0.658823529411765,0.407843137254902}
\definecolor{color3}{rgb}{0.768627450980392,0.305882352941176,0.32156862745098}
\definecolor{color4}{rgb}{0.505882352941176,0.447058823529412,0.698039215686274}
\definecolor{color5}{rgb}{0.8,0.725490196078431,0.454901960784314}

\begin{groupplot}[group style={group size=2 by 1}]
\nextgroupplot[
axis background/.style={fill=color0},
axis line style={white},
legend cell align={left},
legend style={at={(0.97,0.03)}, anchor=south east, draw=none, fill=color0},
tick align=outside,
tick pos=left,
title={Model accuracy (training)},
x grid style={white},
xlabel={epoch},
xmajorgrids,
xmin=-0.45, xmax=9.45,
xtick style={color=white!15.0!black},
y grid style={white},
ylabel={accuracy},
ymajorgrids,
ymin=0.977800427178351, ymax=0.999792724310862,
ytick style={color=white!15.0!black}
]
\addplot [line width=0.7000000000000001pt, color1]
table {%
0 0.981353863909546
1 0.994909922539072
2 0.997376248785222
3 0.997778557304821
4 0.998093407450595
5 0.998268324198247
6 0.998688124392611
7 0.998355782572072
8 0.998408257596368
9 0.998513207613682
};
\addlegendentry{Model 1}
\addplot [line width=0.7000000000000001pt, color2]
table {%
0 0.980671686977695
1 0.994735005478645
2 0.996973940265623
3 0.997866015678647
4 0.99811089912536
5 0.998390765921603
6 0.998163374149656
7 0.998233340848716
8 0.998513207644959
9 0.998513207644959
};
\addlegendentry{Model 2}
\addplot [line width=0.7000000000000001pt, color3]
table {%
0 0.979709644839545
1 0.994157780284375
2 0.997446215484283
3 0.997463707159048
4 0.998320799222542
5 0.998163374149656
6 0.998478224295429
7 0.99814588247489
8 0.998670632717846
9 0.998688124392611
};
\addlegendentry{Model 3}
\addplot [line width=0.7000000000000001pt, color4]
table {%
0 0.97880007704801
1 0.995172297764809
2 0.997148857013275
3 0.997796048979587
4 0.99814588247489
5 0.998128390800125
6 0.998390765869474
7 0.998478224295429
8 0.998740599416907
9 0.998600666018785
};
\addlegendentry{Model 4}
\addplot [line width=0.7000000000000001pt, color5]
table {%
0 0.980269378353838
1 0.994874939189542
2 0.996781531738947
3 0.997813540654352
4 0.998023440647275
5 0.998005949076769
6 0.998233340848716
7 0.998408257596368
8 0.998688124392611
9 0.998793074441202
};
\addlegendentry{Model 5}

\nextgroupplot[
axis background/.style={fill=color0},
axis line style={white},
legend cell align={left},
legend style={draw=none, fill=color0},
tick align=outside,
tick pos=left,
title={Model loss (training)},
x grid style={white},
xlabel={epoch},
xmajorgrids,
xmin=-0.45, xmax=9.45,
xtick style={color=white!15.0!black},
y grid style={white},
ylabel={loss},
ymajorgrids,
ymin=-0.000910058736891878, ymax=0.0750433519511403,
ytick style={color=white!15.0!black}
]
\addplot [line width=0.7000000000000001pt, color1]
table {%
0 0.0644373292752094
1 0.014538030518019
2 0.00860862545817576
3 0.00647704879139864
4 0.0054134360630892
5 0.00445699468298795
6 0.00321765908805707
7 0.00433922619462015
8 0.0038192960786329
9 0.00394761440892392
};
\addlegendentry{Model 1}
\addplot [line width=0.7000000000000001pt, color2]
table {%
0 0.0668259143467658
1 0.015464917693059
2 0.00865809195423193
3 0.00646747387920555
4 0.00541821951020739
5 0.0038347508300218
6 0.00463851704308058
7 0.00442272979805627
8 0.00388066108086871
9 0.00390859629208228
};
\addlegendentry{Model 2}
\addplot [line width=0.7000000000000001pt, color3]
table {%
0 0.0698609500515796
1 0.0171374021390553
2 0.00835407340473584
3 0.00668820448924733
4 0.00514445256928618
5 0.00471206591722036
6 0.00393885065668262
7 0.0041923497455356
8 0.00334336654959967
9 0.00352429966894075
};
\addlegendentry{Model 3}
\addplot [line width=0.7000000000000001pt, color4]
table {%
0 0.0715909241925933
1 0.0151021538091454
2 0.00926834612400229
3 0.00648553751249474
4 0.00526328353206745
5 0.00497001530574402
6 0.00455669408762635
7 0.00386104030565405
8 0.00354370728914077
9 0.00377768975961792
};
\addlegendentry{Model 4}
\addplot [line width=0.7000000000000001pt, color5]
table {%
0 0.0684375002074404
1 0.0155832685019441
2 0.00886865556427625
3 0.00629955474040319
4 0.00542412329568273
5 0.00529344981760299
6 0.00414305358474231
7 0.00353453351094489
8 0.00329702377568452
9 0.00254236902165504
};
\addlegendentry{Model 5}
\end{groupplot}

\end{tikzpicture}
}
\resizebox{8.5cm}{!}{%
\begin{tikzpicture}

\definecolor{color0}{rgb}{0.917647058823529,0.917647058823529,0.949019607843137}
\definecolor{color1}{rgb}{0.298039215686275,0.447058823529412,0.690196078431373}
\definecolor{color2}{rgb}{0.333333333333333,0.658823529411765,0.407843137254902}
\definecolor{color3}{rgb}{0.768627450980392,0.305882352941176,0.32156862745098}
\definecolor{color4}{rgb}{0.505882352941176,0.447058823529412,0.698039215686274}
\definecolor{color5}{rgb}{0.8,0.725490196078431,0.454901960784314}

\begin{groupplot}[group style={group size=2 by 1}]
\nextgroupplot[
axis background/.style={fill=color0},
axis line style={white},
legend cell align={left},
legend style={at={(0.97,0.03)}, anchor=south east, draw=none, fill=color0},
tick align=outside,
tick pos=left,
title={Model accuracy (validation)},
x grid style={white},
xlabel={epoch},
xmajorgrids,
xmin=-0.45, xmax=9.45,
xtick style={color=white!15.0!black},
y grid style={white},
ylabel={accuracy},
ymajorgrids,
ymin=0.993591274564932, ymax=0.999132443281716,
ytick style={color=white!15.0!black}
]
\addplot [line width=0.7000000000000001pt, color1]
table {%
0 0.994332895630562
1 0.997481286946916
2 0.997761143952815
3 0.998320857964611
4 0.998110965210187
5 0.998250893713136
6 0.997971036707238
7 0.998041000958713
8 0.99846078646756
9 0.998180929461662
};
\addlegendentry{Model 1}
\addplot [line width=0.7000000000000001pt, color2]
table {%
0 0.995872109163002
1 0.996711680180696
2 0.997551251198391
3 0.997971036707238
4 0.997971036707238
5 0.998250893713136
6 0.998041000958713
7 0.997901072455764
8 0.997901072455764
9 0.998110965210187
};
\addlegendentry{Model 2}
\addplot [line width=0.7000000000000001pt, color3]
table {%
0 0.99384314587024
1 0.996991537186595
2 0.997621215449865
3 0.997901072455764
4 0.997831108204289
5 0.997551251198391
6 0.998180929461662
7 0.998250893713136
8 0.998810607724933
9 0.998320857964611
};
\addlegendentry{Model 3}
\addplot [line width=0.7000000000000001pt, color4]
table {%
0 0.995102502396782
1 0.997761143952815
2 0.998180929461662
3 0.99846078646756
4 0.997901072455764
5 0.998880571976407
6 0.998600714970509
7 0.998320857964611
8 0.998530750719035
9 0.998670679221984
};
\addlegendentry{Model 4}
\addplot [line width=0.7000000000000001pt, color5]
table {%
0 0.995592252157104
1 0.998041000958713
2 0.99769117970134
3 0.99769117970134
4 0.998110965210187
5 0.998250893713136
6 0.998530750719035
7 0.99846078646756
8 0.998810607724933
9 0.998250893713136
};
\addlegendentry{Model 5}

\nextgroupplot[
axis background/.style={fill=color0},
axis line style={white},
legend cell align={left},
legend style={draw=none, fill=color0},
tick align=outside,
tick pos=left,
title={Model loss (validation)},
x grid style={white},
xlabel={epoch},
xmajorgrids,
xmin=-0.45, xmax=9.45,
xtick style={color=white!15.0!black},
y grid style={white},
ylabel={loss},
ymajorgrids,
ymin=0.00169154729892136, ymax=0.0166343110604796,
ytick style={color=white!15.0!black}
]
\addplot [line width=0.7000000000000001pt, color1]
table {%
0 0.0159550945258633
1 0.00813168665926343
2 0.00549212226945748
3 0.00468638291153566
4 0.00553928926344705
5 0.00389491389020953
6 0.00430044043957615
7 0.00441388887731122
8 0.00298318468576082
9 0.00330635699319081
};
\addlegendentry{Model 1}
\addplot [line width=0.7000000000000001pt, color2]
table {%
0 0.0124475428818463
1 0.00782269231540156
2 0.00559767637670756
3 0.00689270784885637
4 0.00663435028168549
5 0.00497459688651307
6 0.00508672732813372
7 0.00487092428948158
8 0.00440323077459539
9 0.00358813275924878
};
\addlegendentry{Model 2}
\addplot [line width=0.7000000000000001pt, color3]
table {%
0 0.0152258819727194
1 0.00746091385872896
2 0.0055387276551729
3 0.00561946865736447
4 0.00547087302945066
5 0.00469235263817928
6 0.00472570463744195
7 0.00353540861404279
8 0.00237076383353764
9 0.00440610958334518
};
\addlegendentry{Model 3}
\addplot [line width=0.7000000000000001pt, color4]
table {%
0 0.0156107467695747
1 0.00704636887710188
2 0.00493308763679935
3 0.00371309291140175
4 0.00543436256335653
5 0.00391685281317296
6 0.00401969348959342
7 0.00345901310709413
8 0.00355426343035182
9 0.00336351340717229
};
\addlegendentry{Model 4}
\addplot [line width=0.7000000000000001pt, color5]
table {%
0 0.0131551067060163
1 0.00627150312994762
2 0.00572775104013003
3 0.00786534045237755
4 0.00403255961342239
5 0.00412565096346121
6 0.00425867271162701
7 0.00338299520781169
8 0.00290245564648513
9 0.00503544499606107
};
\addlegendentry{Model 5}
\end{groupplot}

\end{tikzpicture}
}
\caption{Accuracies and losses of CNN submodels}
\label{fig:acc_loss_fig1}
\end{figure}

The performance of the submodels and their combination are evaluated on the accuracy of the trained systems. The final results  reported are the average 10-fold cross-validation accuracies. There is a significant performance variation between the five submodels by categories on the validation data. Figure \ref{fig:acc_loss_fig1} shows the curve of accuracies and losses during the training and validation step of the CNN submodels of the treatment category. This regularization effect on validation data is substantially similar on all categories.

 Analysis of the accuracy  results on the validation data shows a convergence stability to $99\%$ obtained with the submodels of Lifestyle and Unknown categories, while about more than half of Consultation subject, personal and medical history submodels have an accuracy that varies between $97\%$ and $99\%$. Those in the symptom category did not exceed $98\%$. 
We note that each of the obtained accuracies is significantly representative of the whole training and validation datasets.
Table \ref{tab:moy_acc_validation} presents a simple average of the submodel accuracies for each category that is used as $\alpha_e$.

\begin{table}[htb]
\centering
\begin{tabular}{@{}l|l|l|@{}}
 & CNN & FastText \\
\rowcolor{Gray}
\begin{tabular}[c]{@{}l@{}}Consultation\\ subject\end{tabular} & 0.98\% & 0.98\% \\
Personal & 0.98\% & 0.99\% \\
\rowcolor{Gray}
\begin{tabular}[c]{@{}l@{}}Medical\\ history\end{tabular} & 0.97\% & 0.99\% \\
Symptoms & 0.97\% & 0.98\% \\
\rowcolor{Gray}
Lifestyle & 0.99\% & 0.99\% \\
Treatment & 0.98\% & 0.98\% \\
\rowcolor{Gray}
Unknow & 0.99\% & 0.99\% \\
\end{tabular}
\caption{Average accuracies per category on the validation data}
\label{tab:moy_acc_validation}
\end{table}

We then evaluated the two classification systems and their combination on the test data from the test sessions with the interns in medicine typically dealing with the clinical case studied. Table \ref{tab:test_accuracy} shows the test accuracies of different systems averaged over the 10 folds. The classifier combination is performed according to algorithm \ref{algo:a1}.

\begin{table}[htb]
\centering
\begin{tabular}{@{}lll@{}}
\toprule
CNN & FastText & Combined \\ \midrule
68.09\% & 94.78\% & {\bf 96.80\%} \\
\midrule
\end{tabular}
\caption{Accuracy of classification systems on test sessions data}
\label{tab:test_accuracy}
\end{table}

For single classification systems, FastText performs a lot better than CNN. But the combination allowed to gain about $2\%$ performance compared to the FastText based system.

\subsection{Similar Question Identification}
\label{sec:sqi}

Identification of similar questions is performed by semantic comparison between the distributed representations of the two sentences. The goal is to extract from the virtual patient knowledge base, a question semantically close to the question asked by the student. Knowledge is organized by category of questions in order to limit the scope of the semantic comparison. This has therefore enabled us to infer more intent questions by improving the completeness of studied clinical case. Table \ref{tabquestion_reformulation} shows some examples of questions and knowledge obtained by calculating semantic distances.

\begin{table*}[htb]
\centering
\begin{tabular}{@{}clll@{}}
\toprule
\begin{tabular}[c]{@{}l@{}}\textbf{Question} \\ \textbf{category}\end{tabular} & \textbf{Asked question} & \textbf{Knowledge base} & \textbf{Expected response} \\ \midrule
\multirow{5}{*}{Symptom} & do you have temperature? & Are you feverish? & No, I have no fever. \\ 
 & \cellcolor{Gray} Where is your pain located? & \cellcolor{Gray} where exactly do you hurt? & \cellcolor{Gray} I hurt in the stomach. \\ 
 & Is the pain is in the stomach? & where exactly do you hurt? & I hurt in the stomach. \\ 
 & \cellcolor{Gray} Do you have dark urine? & \cellcolor{Gray} what is the color of your urine? & \cellcolor{Gray} I have clear urine \\ 
 & How is your pee? & what is the color of your urine? & I have clear urine \\ \midrule
\multirow{4}{*}{\begin{tabular}[c]{@{}c@{}}Medical \\ history \end{tabular}} & \cellcolor{Gray}
 Do you have any health problem? & \cellcolor{Gray} what is your medical history? & \cellcolor{Gray} No, nothing special. \\ 
 & Have you suffered an illness in your past? & what is your medical history? & No, nothing special. \\ 
 & \cellcolor{Gray} A particular problem in your family? & \cellcolor{Gray} Do you have a family history? &\cellcolor{Gray}  No doctor \\ 
 & Is there a known disease in your environment? & Do you have a family history? & No doctor \\ \bottomrule
\end{tabular}
\caption{Some similar questions found by calculating semantic distance}
\label{tabquestion_reformulation}
\end{table*}

It consists essentially of a calculator of cosine distances between vectors of sentence embeddings. The semantic distance between questions $q_i$ and $q_j$ is calculated from the word vectors $(w_{i1}, ..., w_{in})$ and $(w'_{j1}, ..., w'_{jn})$ as in Equation \ref{eq:cos}:

\begin{equation}
        dist (q_i, q_j) =  1  - 
        cos\left(\sum_{k=0}^{N}\sigma_k v(w_{ik});\sum_{k'=0}^{N'}\sigma'_{k'} v(w'_{jk'})\right)
        \label{eq:cos}
\end{equation}
    
where $\sigma_k$  and $\sigma'_{k'}$ are respectively the weighting of the words $k$ and $k'$ with their inverse frequencies in documents.

\section{Improving Dialog System}
\label{sec:ids}

Both the developed dialogue systems were deployed at the faculty of medicine for testing sessions with the students. The performance results are reported in section \ref{sec:pa}. In order to improve the understanding of the virtual patient and make the system usable in terms of educational objectives, we combined the two systems in a single dialogue manager. The integrated system therefore consists of the pattern matching system and the system using textual semantic similarity for purposes of good resilience of the dialogue with the user. The combination is made to allow the virtual patient to provide the answer to a question for which the system has no matching rule.

\begin{figure}[htb]
\centering
\resizebox{9cm}{!}{%
\begin{tikzpicture}[font=\sffamily]
\tikzstyle{element}=[rectangle,draw,fill=white,rounded corners=4pt]
\tikzstyle{terminal}=[circle,draw]
\tikzstyle{arrow}=[->,>=stealth,thick,rounded corners=4pt]

\draw[] (-3,0.9) -- (13.2,0.9);

\node[] (pms) at (4,0.6) {PATTERN MATCHING SYSTEM};
\node[] (question) at (-1,0) {\Large user question};
\node[element] (finder) at (4,0) {\Large rule pattern finder};
\node[] (response) at (10.5,0) {\Large indicated response};
\draw[arrow] (question) -- (finder); 
\draw[arrow] (finder) -- (response) node[midway,above,scale=0.7] {$Yes$};

\draw[arrow] (finder) -- (4,-1) node[midway,right,scale=0.7] {$No$};

\draw[] (-3,-1) -- (13.2,-1);

\node[] (ssbs) at (4,-1.3) {SEMANTIC SIMILARITY BASED SYSTEM};
\node[element,text width=3cm,align=center] (categorizer) at (-1,-2) {\Large question\\ categorization};
\node[element] (distancer) at (3.5,-2) {\Large distance calculation};
\node[element,text width=3.5cm,align=center] (closekfinder) at (8,-2) {\Large close knowledge\\finder};
\node[below right,text width=3cm,align=center] (rephrase) at (10.5,-1.75) {\Large ask to rephrase\\ the question};

\draw[arrow] (categorizer) -- (distancer); 
\draw[arrow] (distancer) -- (closekfinder);
\draw[arrow] (closekfinder.10) -| (response) node[near start,above,scale=0.7] {$Yes$};
\draw[arrow] (closekfinder.-10) -- (rephrase) node[near start,below,scale=0.7] {$Yes$};

\draw[] (-3,-2.8) -- (13.2,-2.8);

\end{tikzpicture}
}
\caption{Combined system for improving the dialog manager}
\label{fig:combinedsystem}
\end{figure}
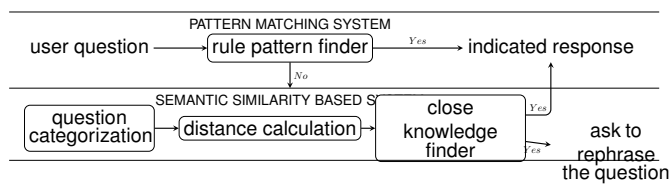

The process of input question interpretation is described in Figure \ref{fig:combinedsystem} which illustrates each processing step in the combined system. When a question is asked to the virtual patient, the manager first looks for a rule pattern corresponding to the input sentence linguistic structure. The rule patterns are kinds of regular expressions whose atoms are words and concepts that group together word equivalence classes. If the rule pattern is found, the corresponding response is provided to the user. Otherwise, the manager sends the normalized form of the question to each binary classifiers for identifying its category. Then, depending on the category, the manager looks for a semantically close question in its knowledge base by calculating cosine distance between the asked question and each question from the category in its knowledge base. According to the answer found, a response is provided to the user.

\section{Performance Analysis}
\label{sec:pa}

We conducted a comparative analysis of performance between the dialogue system using ChatScript only and the systems based on our proposed combination approach whose hypothesizes an improvement of question understanding and interpretation. The evaluation is performed with two different datasets. A first dataset is issued from sessions with students who, before the tests, were briefed on the patient's situation and whose questions were perfectly in line with the case of a surgical consultation of abdominal pain. The second dataset is extracted from \textit{Chicago Med} television series dialogues. It contains questions that may differ from the context of the studied case. The latter dataset is expected to be more difficult for the system.
Please note that this evaluation is not intended to evaluate the student's overall performance in virtual patient interview but to evaluate the virtual patient ability to provide coherent responses to the student. The pedagogical evaluation of the full system including the virtual reality aspects is currently ongoing.

For the current evaluation, we defined three quantitative measures used to compute precision, recall and $F1$ scores for each system:

\begin{itemize}
    \item \textbf{CRR}: the Coherent Responses Rate, measuring questions understanding;
    \item \textbf{IRR}: the Incoherent Responses Rate, measuring  questions misunderstanding;
    \item \textbf{NRR}: the Not recognized Responses Rate (\textit{''I did not understand''}), measuring questions that are identified as not understood.
\end{itemize}

The answers given by the system to the student are annotated with these categories by two doctor experts.

Precision is defined as the total number of coherent responses divided by the total number of coherent and incoherent responses. Recall is defined as the total number of coherent responses divided by the total number of known questions. Finally, the overall performance of the systems is measured with $F1$-score that allows to seek a balance between Precision and Recall.

\begin{table*}[htb]
\centering
\begin{tabular}{@{}lllllll@{}}
\toprule
\textbf{System based on} & \textbf{CRR} & \textbf{IRR} & \textbf{NRR} & \textbf{Precision} & \textbf{Recall} & \textbf{F1-score} \\ \midrule
\rowcolor{Gray}
Rules & 87.81 & \textbf{2.49} & 9.70 & \textbf{97.24} & 87.81 & 92.29 \\
Rules+CNN & 93.18 & 5.68 & 1.13  & 94.25 & 93.18 & 93.71 \\
\rowcolor{Gray}
Rules+FastText  & 94.31 & 4.82 & 0.85 & 95.12 & 94.31 & 94.72 \\
Rules+CNN+FastText  & \textbf{94.88} & 5.11 & \textbf{0} & 94.88 & \textbf{94.88} & \textbf{94.88} \\ \bottomrule
\end{tabular}
\caption{Evaluation of systems on data from test sessions}
\label{tab:results_dataset1}
\end{table*}

Table \ref{tab:results_dataset1} shows the results of the evaluation on the questions from the test sessions with the students. We can notice a growth of the $F1$-score going from the simple Rules-Based system to the system combining rules, CNN and FastText classifiers and the semantic similarity (full system).  The improvement gained with the full system is significant ($ 2.58 \% $ of $F1$-score) in the sense that we obtained an increase in the coherent responses ($\simeq 7\%$ of CRR) and a reduction of non-understanding questions ($\simeq 10\%$ of NRR). There is no significant improvement in performance between the Rules+CNN and Rules+FastText systems despite the overall accuracy gained by FastText in the classification phase on the test data. When the combined systems, at the beginning of the process, don't find rules for a question, the semantic similarity module extracts sometimes incoherent responses from the knowledge base. This is materialized by the increase in the incoherent response score ($\simeq 3\%$ of IRR) observed between the rule-based system and the different combined systems. Thus we can notice that by reducing the number of non-understanding questions, the number of incoherent responses increases even if it is at a low rate. Figure \ref{fig:ncr_nir_nnr} clearly shows the evolution shape of CRR, IRR and NRR.
This increase in incoherent responses does not impede the patient interview or the ability to obtain the right diagnosis for the student doctor because it happens in real cases that the patient does not always understand the question asked by the doctor. The gain is the significant improvement in the number of non-understanding questions associated to the increase of coherent responses. This gain reduces frustration and disappointment of the student often caused by the virtual patient \textit{"I did not understand"} responses, leading the student to feel that the system is not useful at all.

\pgfplotstableread{questions_understanding.txt}{\questionsunderstanding}

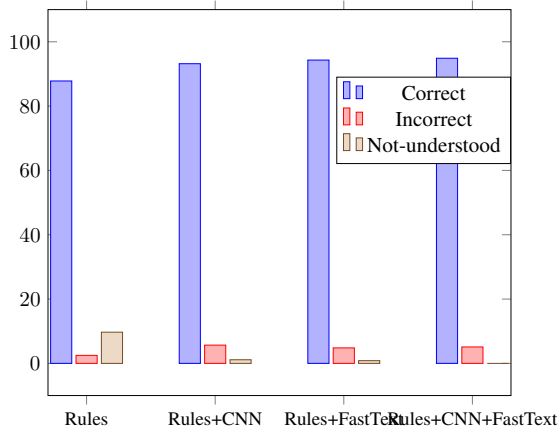
\begin{figure}[htb]
\centering
\begin{tikzpicture}[scale=0.8]

%
	
    \begin{axis}[
		legend entries={Correct, Incorrect, Not-understood},
		legend style={at={(1,0.6)},anchor=south east},
        height=8cm,
        ybar,
        ymin = 0,
		ymax = 100,
        log ticks with fixed point,
        enlargelimits=0.1,
        xtick=data,
        xticklabels from table={\questionsunderstanding}{label},
        x tick label style={anchor=north,font=\small}
    ]
        \foreach \i in {CRR,IRR,NRR}
        {\addplot table [x expr=\coordindex, y=\i] {\questionsunderstanding}; }
    \end{axis}
\end{tikzpicture}

\caption{System responses quality evaluated on data from test sessions. 
}
\label{fig:ncr_nir_nnr}
\end{figure}

Table \ref{tab:results_dataset2} shows the evaluation results on 
\textit{Chicago Med} data which includes questions encountered during a medical consultation not dealing with the clinical case.  We obtained an overall performance ($F1 = 92.15\%$) with a gain of $\simeq 10\%$ on Rules+CNN+FastText system. The NRR score ($=27.72\%$) of Rules-Based system demonstrates the gap between \textit{Chicago Med} questions and the clinical case questions on which we focused the writing of rule patterns. Our proposed approach increases the overall performance of the dialogue system while reducing the number of non-understanding questions.
The resulting gain is significant which shows that without adding new rules, question categorization and semantic similarity effectively complete the system based only on rules and make the combined system more efficient. 

\begin{table*}[htb]
\centering
\begin{tabular}{@{}lllllll@{}}
\toprule
\textbf{System based on} & \textbf{CRR} & \textbf{IRR} & \textbf{NRR} & \textbf{Precision} & \textbf{Recall} & \textbf{F1-score} \\ \midrule
\rowcolor{Gray}
Rules & 70.79 & \textbf{1.49} & 27.72 & \textbf{97.95} & 70.79 & 82.18 \\
Rules+CNN & 85.64 & 7.92 & 3.44 & 91.53 & 85.64 & 88.49 \\
\rowcolor{Gray}Rules+FastText & 88.12 & 4.95 & 6.93 & 94.68 & 88.12 & 91.28 \\
Rules+CNN+FastText & \textbf{90.10} & 5.45 & \textbf{4.46} & 94.30 & \textbf{90.10} & \textbf{92.15} \\ \bottomrule
\end{tabular}
\caption{Evaluation of systems on data from \textit{Chicago Med}}
\label{tab:results_dataset2}

\end{table*}

Although the tasks are slightly different, we compare our work to \cite{jin2017} because we share the same assumption of providing more relevant information using the Convolutional Neural networks combined with a rule-based system for  question identification. The methods are different in the sense that we use a semantic similarity function based on pre-trained word embeddings for the selection of related questions that reduced the questions identified as not understood to $0\%$. 
Compared to \cite{jin2017}, we used in addition to CNNs, a second classification model with a decision-making mechanism for the question identification task n a virtual patient dialogue system. This has impressively improved the identification of  similar questions for an understanding by the virtual patient with overall performance to $94.88\%$.

\section{Conclusion}
\label{sec:conclusion}

In this paper, we described a voice-based dialogue system for the medical student training in the diagnosis of surgical emergencies.
Our system combines the description capabilities of a dialogue scenario with pattern-based rules to the resilience provided by semantic similarity based on word embeddings. With the combined FastText and CNN models, the system significantly improves its performance compared to the versions using the rules and either the CNN submodel or the FastText submodel. With this overall performance, we achieved an understanding rate that makes the system usable. We thus obtain a conversational virtual standardized patient system that allows medical students to simulate a diagnostic strategy of an abdominal surgical emergency.

We are currently preparing the pedagogical evaluation of the full system which integrates the dialogue system inside a realistic virtual reality scene. This system includes the filling of the database summarizing the patient and disease models. This data allows to pedagogically evaluate the quality of the student work and the impact on the doctors, which are the ultimate goal of this project. 

\bibliographystyle{IEEEtran}
\bibliography{main} 

\end{document}